\documentclass[sigconf,authorversion,leftjustify]{acmart}

\renewcommand\footnotetextcopyrightpermission[1]{} 
\acmDOI{}
\acmConference[]{}{}{}%
\makeatletter
\renewcommand\@formatdoi[1]{\ignorespaces}
\setcopyright{none}
\usepackage{times}
\usepackage{epsfig}
\usepackage{graphicx}
\usepackage{tabularx}
\usepackage{tablefootnote}
\usepackage{subfigure}
\usepackage{booktabs}
\usepackage{tabularray} 
\usepackage{extpfeil}
\usepackage{xcolor}
\usepackage{hyperref}
\hypersetup{
    colorlinks=true,
    citecolor=red,
    linkcolor=magenta,
    filecolor=blue,      
    urlcolor=blue,
    pdftitle={Overleaf Example},
    pdfpagemode=FullScreen,
    }

\usepackage{amsmath}
\usepackage{amssymb}
\begin{document}

\title{Extracting Self-Consistent Causal Insights from Users Feedback with LLMs and In-context Learning}
\author{Sara Abdali}
\authornote{sabda005@ucr.edu}
\email{saraabdali@microsoft.com}
\affiliation{%
  \institution{Microsoft Corporation}
  \city{Redmond}
  \state{Washington}
  \country{USA}
}
\author {Anjali Parikh}
\email{anjalip@microsoft.com}
\affiliation{%
  \institution{Microsoft Corporation}
  \city{Redmond}
  \state{Washington}
  \country{USA}
}

\author {Steve Lim}
\email{steve.lim@microsoft.com}
\affiliation{%
  \institution{Microsoft Corporation}
  \city{Redmond}
  \state{Washington}
  \country{USA}
}
\author{Emre Kiciman}
\email{emrek@microsoft.com}
\affiliation{%
  \institution{Microsoft Research}
  \city{Redmond}
  \state{Washington}
  \country{USA}
}
\begin{abstract}
Microsoft Windows Feedback Hub is designed to receive customer feedback on a wide variety of subjects including critical topics such as power and battery. Feedback is one of the most effective ways to have a grasp of users' experience with Windows and its ecosystem. However, the sheer volume of feedback received by Feedback Hub makes it immensely challenging to diagnose the actual cause of reported issues. To better understand and triage issues, we leverage Double Machine Learning (DML) to associate users' feedback with telemetry signals. One of the main challenges we face in the DML pipeline is the necessity of domain knowledge for model design (e.g., causal graph), which sometimes is either not available or hard to obtain. In this work, we take advantage of reasoning capabilities in Large Language Models (LLMs) to generate a prior model that which to some extent compensates for the lack of domain knowledge and could be used as a heuristic for measuring feedback informativeness. Our LLM-based approach is able to extract previously known issues, uncover new bugs, and identify sequences of events that lead to a bug, while minimizing out-of-domain outputs.   
\end{abstract}
\keywords{Large Language Models, Causal Inference, Double Machine Learning, In Context Learning, Prompt Engineering,Natural Language Inference}

\maketitle

\section{Introduction}
 Feedback Hub is a Windows application produced by Microsoft. It allows general Windows users and Windows Insider users to provide feedback, feature suggestions, and bug reports for the Windows operating system and connected devices. Feedback Hub receives feedback on a wide range of subjects, including issues associated with power and battery. 
  \par Triaging reported bugs usually involves categorizing issues, pulling requests based on priority or urgency, and then associating them with telemetry signals for causal analysis. Triage is often a tedious and time-consuming task for Microsoft engineers. Therefore, Microsoft researchers and data scientists have been trying to automatize some of the triage modules such as classification, summariazation, and topic modeling. However, there are still areas such as the causal inference pipeline that highly relies on domain knowledge. 
  \par Microsoft has been always prioritizing causal inference initiatives by developing publicly available tools like EconML\footnote{https://github.com/py-why/EconML}~\cite{kiciman2022a} and DoWhy~\footnote{https://github.com/py-why/dowhy}~\cite{Sharma2020DoWhyAE} libraries. These libraries are among the most efficient existing tools for real world end-to-end causal analysis. However, as mentioned, domain knowledge is still an inevitable necessity to take full advantage of them, specially for casual modeling and graph construction. 
  \par In recent years, generative Artificial Intelligence (AI) has revolutionized different areas of research including Double Machine Learning (DML). Researchers have been using generative models such as Variational Auto Encoders (VAE) to produce priors for causal modeling~\cite{Geffner2022DeepEC} and more recently Large Language Models (LLMs) have been exploited for causal reasoning~\cite{Kcman2023CausalRA,long2023large}. Although LLMs achieve astonishing results in causal reasoning tasks, they are susceptible to hallucination and incorrect answers.
  \par In this work, we leverage In-Context Learning (ICL) to design a modified self-consistency framework to mitigate LLMs' hallucination to extract reliable causal variables (e.g., treatment, outcome, confounders) from users feedback in order to generate a prior causal model and semi-automatize the graph construction process in the causal pipeline. Moreover, we extract chains of events from feedback to provide engineers with a "causal summary" of reported bugs. Finally, we will show how we can leverage the causal model and sequences to score feedback actionability.
  Overall, our contributions are as follows:
  \begin{itemize}
      \item We propose a a modified self-consistency approach which leverages an ensemble of prompts and increases the chance of unconfoundedness via a greedy approach. 
      \item We take advantage of reasoning capabilities of LLMs and ICL to extract causal variables and sequences of events from Feedback Hub.
      \item We Design two casual heuristics for scoring information richness of feedback. 
  \end{itemize}
\begin{figure*}
  \includegraphics[width=\linewidth]{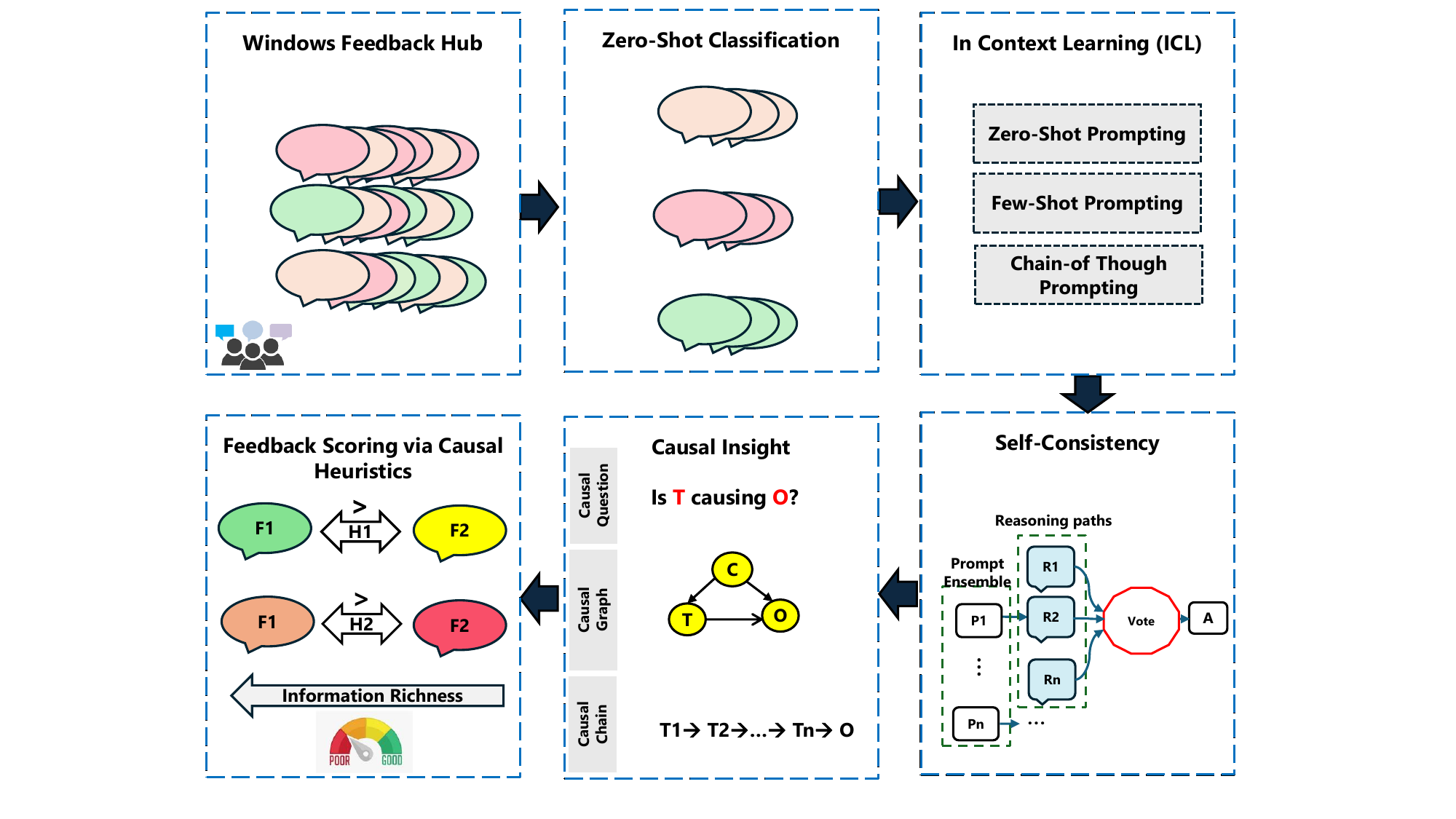}
  \caption{Overview of the proposed approach.}
  \label{fig:overview}
\end{figure*}
\section{Background}
\subsection{Causal Inference Pipeline}
Every time we try to answer a question that asks \textit{"Why?"}, we are engaged in causal analysis.
In fact, we attempt to determine a \textit{cause} for an observed \textit{outcome} which has occurred due to an \textit{action}. In other words,
Causal Inference is the process of inferring causes from the data. To estimate the effect, the gold standard is to conduct a randomized experiment where a randomized subset of units is acted upon and the other subset is not. For instance, Average Treatment Effect (ATE) of a binary treatment is as follows~\cite{imbens_rubin_2015}:
\begin{equation}
    \text{ATE}=E(Y|T=1)-E(Y|T=0)
\end{equation} 
However, as it is not always feasible to run a randomized experiment, we often resort to observational or logged data. In the case of bugs that exist within the OS due to the interaction of the operating system, attached devices, drivers loaded, applications used, and use of cloud-based services, it becomes impossible to diagnose issues that could be from the interaction of many attributes.
Such observed data are susceptible to bias by correlations and unobserved common causes. Causal inference aims to identify and mitigate such biases. A causal inference study usually has the following steps~\cite{Sharma2020DoWhyAE,kiciman2022a}: 
\begin{itemize}
    \item \textbf{Modeling:} The first step in causal inference pipeline is the modeling step, where we encode our domain knowledge into a causal model, often represented as a graph~\cite{pearl_2009}. Other than \textit{Treatment} and \textit{Outcome}, we may identify the following:
    \begin{itemize}
        \item \textbf{Confounders or Common Causes}: These are variables that cause both the action and the outcome. As a result, any observed correlation between the action and the outcome may simply be due to the confounding variables, and not due to any causal relationship between them.
        \item  \textbf{Instrumental Variables:} These are special variables that cause the action, but do not directly affect the outcome. They are not affected by any variable that affects the outcome. If there is an instrumental variable available, then we can estimate effect even if any (or none) of the common causes of treatment and outcome are unobserved.
~\cite{Syrgkanis2019MachineLE}.
         \item \textbf{Effect Modifiers:}
       These are variables that cause the outcome, but do not directly affect the action. 
    \end{itemize}
The aforementioned variables are cornerstones of a causal graph and identifying them usually requires \textit{domain knowledge}.
  \item \textbf{Identification:} Checking if the target quantity (e.g., effect of A on B) can be estimated given the observed variables. Causal estimands such as \textit{Back-door}, \textit{Front-door}, \textit{Mediation} or \textit{Instrumental Variable } estimands are identified here.
     \item \textbf{Estimation:} building a statistical estimator that computes the target estimand identified in the previous step.
     \item \textbf{Refutation:} One of the most important steps in causal inference, where we Check robustness of the estimate and reject the hypothesis if estimation is sensitive to refutation tests.
\end{itemize}
In this work, we focus on the first step, i.e., "Modeling"  which is arguably the most important step in the causal inference pipeline.

\subsection{Large Language Models and In Context Learning}
In-Context Learning (ICL) is the simplest and one of the most efficient paradigms in Natural Language Understanding (NLU), where a pre-trained LLM is prompted with some instructions or demonstrations (e.g., examples to solve a new task without any training or fine-tuning). For instance, a Causal Language Model (CLM) is conditioned on the prompt to predict the next tokens:
\begin{equation}
    \mathrm{p}(1:n)=\prod_{i=1}^{n} \mathrm{p}(t_i|t_1,\cdots,t_{i-1},\text{prompt},\text{query})
\end{equation}
ICL can be improved by dynamically retrieving demonstrations that are similar to the query input. Some of the widely used in-context prompting techniques are as follows:
\begin{itemize}
    \item \textbf{Zero-Shot Prompting} enables a language model to make predictions about unseen data needless to use any additional training or examples~\cite{wei2022finetuned}.
    \item \textbf{Few-Shot Prompting} a prompting technique where a language model processes in-domain examples before generation. This method was popularized by the original GPT-3 paper~\cite{brown2020language}, where it  is shown to be an emergent property of LLMs when they are scaled to a sufficient size~\cite{Kaplan2020ScalingLF,Touvron2023LLaMAOA}.
    \item \textbf{Chain-of-Though (CoT) Prompting} a prompting technique where a series of intermediate reasoning steps a.k.a. Chain-of-Thoughts (CoT) demonstrations are provided as exemplars in prompting to invoke the ability of large language models to perform complex reasoning tasks. Similar to few-shot capability, it is shown that reasoning abilities emerge naturally in sufficiently large language models via this prompting technique~\cite{wei2022chain}. 
\end{itemize}
It is worth mentioning that CoT is commonly merged with the other two approaches to generate more efficient prompts which result in more accurate responses.

\section{Proposed Framework}
In this section, we discuss our proposed framework for extracting causal insights, and creating causal graph as a prior model for the first step in causal inference pipeline. 
\subsection{Zero-shot Feedback Classification} As mentioned earlier, Feedback Hub receives an overwhelming amount of feedback on a variety of topics in Windows and its ecosystem. Therefore, categorizing feedback into classes of issues is a crucial pre-processing step for any statistical analysis. Thus, we leverage a pre-trained LLM to classify feedback in a zero-shot setting. More specifically, we use LLM in a Natural Language Inference (NLI)~\cite{bowman-etal-2015-large} paradigm by considering the \textit{premise} as a given feedback we would like to classify and the hypothesis as a sentence with the following structure: 
\begin{align*}
    \textit{The topic of this feedback is } \{ \textit{label-name} \}
\end{align*}

where \textit{label-name} is the class of the issue we aim to predict.
Probabilities for entailment, contradiction and neutral tell us whether or not the feedback is classified as the provided label.
\subsection{Extracting Causal Insights Using In-Context Learning}
After categorizing feedback, we leverage ICL to design multiple prompts for the following purposes:
\begin{itemize}
    \item Extracting causal variables e.g., treatment, effect, and confounding variables. 
    \item Extracting Sequences (chains) of events in order to unearth the root cause of reported issue if is mentioned.
\end{itemize}
In the next step, we utilize self-consistency on top of an ensemble of prompts to mitigate hallucination and incorrect reasoning. 
\subsection{Hallucination Mitigation via Self-Consistency and Prompt Ensemble}
Chain-of-Thought prompting combined with few-shot learning has achieves encouraging results on complex reasoning tasks~\cite{wei2022chain}. However, as CoT takes a naive greedy decoding approach, it is susceptible to mistakes and hallucination. To mitigate such mistakes, Wang et al. proposed a new prompting approach a.k.a. self-consistency prompting~\cite{wang2023selfconsistency}
which samples a diverse set of reasoning paths instead of only taking the greedy one, and then selects the most consistent answer by marginalizing out the sampled reasoning paths. The rationale behind this approach is straightforward: \textit{a complex reasoning problem typically admits multiple different ways of thinking leading to its unique correct answer} 
~\cite{wang2023selfconsistency}.
\par In this work, we adopt this approach while making two modifications: 1) We apply self-consistency on an ensemble of prompts designed in the previous step, and 2) After taking the majority vote on the generated treatments and outcomes, we follow a greedy approach by taking union of generated confounders of the elected treatment/outcome pairs in order to maximize the chance of confoundedness. We preserve the study and assessment of this greedy approach for the future~\cite{Unconfoundedness}.
\par In order to generate different reasoning paths, we sample a set of candidate outputs from the language model’s decoder to generate a diverse set of
candidate reasoning paths. These techniques include but not limited to temperature sampling~\cite{ficler-goldberg-2017-controlling}), top-k sampling~\cite{Radford2019LanguageMA}, and nucleus sampling~\cite{DBLP:journals/corr/abs-1904-09751}. In this work, we leverage temperature sampling to generate reasoning paths.
\begin{table}[]
\scriptsize
    \begin{tblr}
{ 
colspec ={p{6cm}|p{1.3cm}},
#row{1} = {lightgray},
}
        \hline
        \textbf{\small Extracted Features}&\textbf{\small Rate}\\
        \hline
         Preciously Known Outcomes(Bugs)/Treatment & 80.9\%\\

          Preciously Unknown Outcomes(Bugs)/Treatment &19.1\% \\
         In-domain Confounders Not Mentioned in Text/Examples &35.7\%\\
         In-domain Confounders Mentioned in Text/Examples&64.3\% \\
         Out-of-Domain Variables (Hallucination) & 0\% \\
        \hline
        Preciously Known Sequences &44\%\\
        Preciously Unknown Sequences&56\%\\
        Out-of-Domain Sequences (Hallucination) & 0\% \\
        \hline
    \end{tblr}
    \caption{Percentage of extracted causal features.}
    \label{table:stat}
\end{table}

\subsection{Causal Modeling and Graph Construction}
In this step, we leverage causal variables extracted in the previous step to create a causal graph. As causal models are Directed Acyclic Graphs (DAG), we discard an edge if it creates a cycle in the graph. Time complexity of cycle detection is of the order of $O(E+V)$, where $E$ and $V$ are number of edges and nodes (variables) respectively.
\subsection{Causal Heuristics for Scoring Feedback}
Not only do the generated insights provide us with required variables for creating causal graphs, but also they could be used to
score actionability of feedback as well.
This is extremely useful, as it helps engineers to reduce the volume
of low quality feedback and prioritize information-rich ones. To this end, we propose:
\begin{itemize}
    \item \textbf{Heuristic 1: Number of Extracted Variables} The higher the number of extracted variables, the richer the feedback.

\item \textbf{Heuristic 2:
Length of the Causal Chain} The longer the length of the event sequence, the richer the feedback.
\end{itemize}
We may use these heuristics solely or in combination to score incoming feedback and filter out less informative ones.
\section{Experiments}
It is typically very hard to obtain ground-truth counterfactual data, as many of the causal variables might not be present in the observational data. In case of Feedback Hub, this becomes even more challenging because users usually do not have enough domain knowledge about the issues they are facing, thereby they report bugs to understand and fix the underlying causes. Moreover, in many cases, users report unknown or emerging issues which are previously unknown to the engineers. Therefore, as we are working on real data instead of synthetic or public datasets, we can either evaluate the results by considering whether or not the proposed method is capable of extracting predefined issues or measure methods capability in extracting unobserved insights. 

\subsection{Dataset and Experimental Settings}
As Feedback Hub receives a massive amount of feedback, we focused on the power and battery related issues to narrow down the scope of our experiments as they have one of the highest priorities in customer satisfaction. To this end, we extracted more than $5000$ users' feedback and applied a topic modeling approach to identify common categories of issues. Then we leveraged a NLI approach as discussed earlier, to classify feedback into one of the 25 subcategories we identified via topic modeling. "Modern Standby" subcategory with roughly $8\%$ prevalence was one of the frequently reported issues. Due to the space limitation, we only report extracted insights of this subcategory. For causal insight extraction, we leveraged GPT-3.5 Turbo model, as it is fast and achieves substantially higher accuracy in reasoning tasks compared to other models of GPT-3 family~\cite{Kcman2023CausalRA}.
\subsection{Experimental Results}
\begin{figure*}
     \centering
     \begin{subfigure}[Zero-shot Chain-of-Though in a self-consistency paradigm]
{\includegraphics[width=1\linewidth]{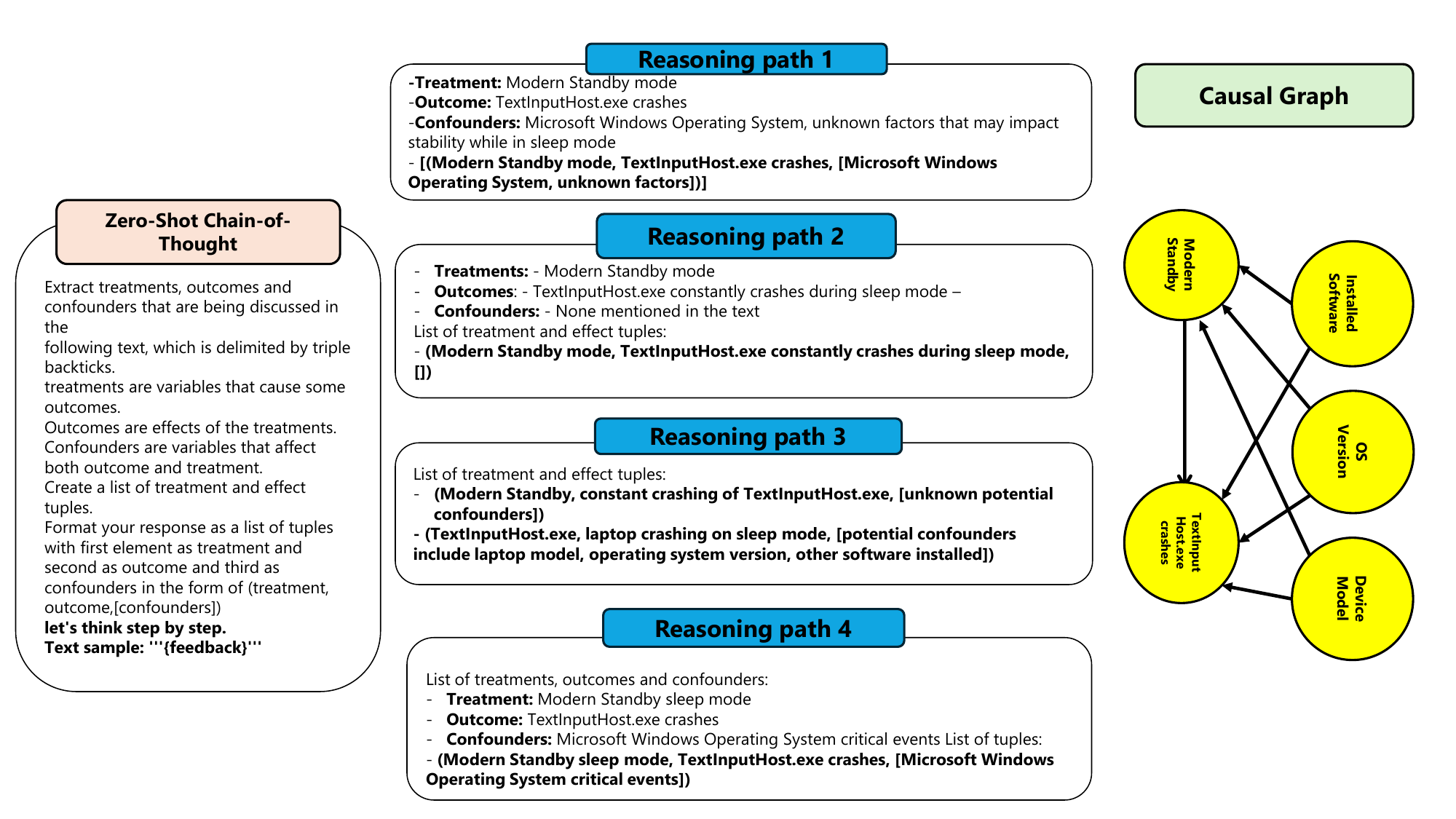}}
     \end{subfigure}
     \begin{subfigure}[Few-shot Chain-of-Though in a self-consistency paradigm]{\includegraphics[width=1\linewidth]{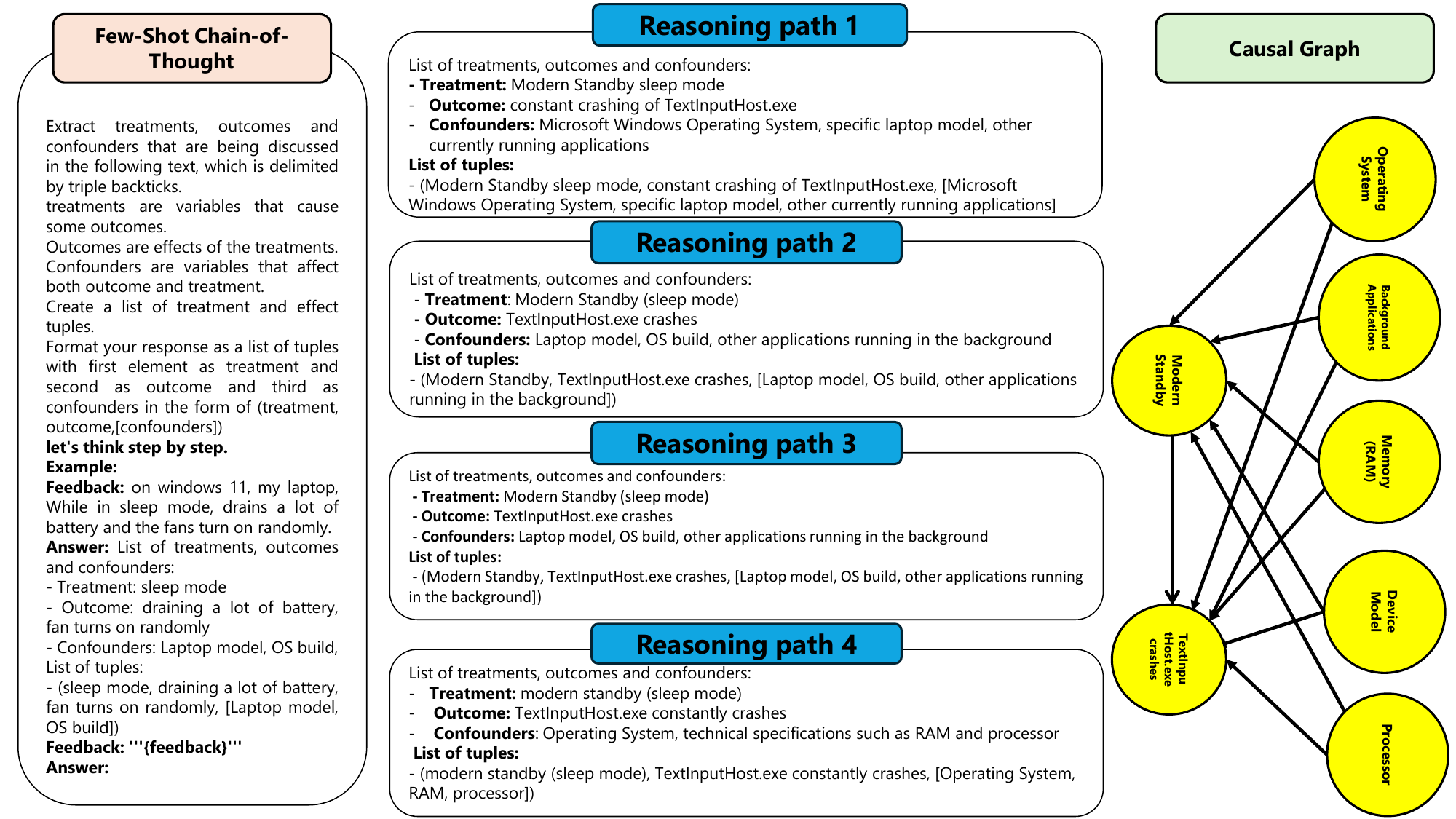}}

     \end{subfigure}
        \caption{Constructing causal graphs with self-consistent Zero-shot and few-shot Chin-of-Thought Prompting.}
        \label{fig:zero/few}
\end{figure*}

We observe that GPT-3.5 Turbo prompted with both zero-shot and few-shot Chain-of-Thought (CoT) prompting is capable of extracting confounders even if they are not mentioned in the text explicitly. This \textit{emergent capability} is specially observed when we slightly increase the temperature of model from zero. However, few-shot CoT provides us with a richer set of variables. Therefore, we used an ensemble of few-shot CoT prompts to extract causal variables via self-consistency approach we discussed earlier. Figure \ref{fig:zero/few}, demonstrates an example of zero-shot and few-shot CoT prompting for a given feedback as well as generated causal graph in our proposed self-consistency paradigm.
\par Table \ref{table:graph} illustrates extracted causal variables and feedback scores using heuristic $1$ (i.e., number of extracted variables). Due to space limitation, we skip graph representations, as having causal variables in hand, graph construction would be straightforward. As denoted in table \ref{table:stat}, around $81\%$ of extracted outcomes/treatments are among bugs/reasons that are pre-classified by engineers. However, $19\%$ of variables are new bugs/treatments that are discovered by model. Another interesting observation is that, $35.7\%$ of extracted variables are merely inferred by LLMs, although they have not been mentioned in the feedback or examples explicitly. This is very promising as this happens while hallucination rate is minimized and none of the variables are identified as out-of domain responses (0\% hallucination).
\par  Table \ref{table:chain} depicts extracted chains of events that result in an outcome (bug) as well as feedback scores using heuristic 2 (i.e., cumulative sequence length). As demonstrated, often more than one chain is extracted from highly detailed feedback which in turn results in higher actionability score. As far as chain of events is concerned, only $44\%$ of chains are previously known by the engineers and the rest of sequences are new chains that are discovered by engineers and worthy of more investigations. Similar to previous case, our self-consistency paradigm has reduced the hallucination rate to $0\%$.

\begin{table*}
\centering
\scriptsize
\begin{tblr}
{ 
colspec ={|p{1.4cm}|p{7cm}|p{7cm}|p{1.1cm}|},
row{1} = {lightgray},
}
\hline
\textbf{Treatment}&\textbf{Outcome}&\textbf{Confounders}&\textbf{Heuristic 1}\\

\hline

\colorbox{blue!23}{Modern Standby}& \colorbox{green!25}{Long Boot up Time}& \colorbox{lime}{Laptop Model},\colorbox{lime}{ BIOS Settings}&4\\
\hline

\colorbox{blue!23}{Modern Standby}& \colorbox{blue!23}{Battery drain}&\colorbox{lime}{Laptop Model}, \colorbox{lime}{OS Version}&4\\
\hline
\colorbox{blue!23}{Modern Standby}& \colorbox{green!25}{Laptop Wakes Up to Carry Out Backup Tasks with Task Scheduler}& \colorbox{green!25}{Task Scheduler Settings}, \colorbox{lime}{Laptop Model}, \colorbox{lime}{OS Build}&5\\

\hline
\colorbox{blue!23}{Modern Standby}& \colorbox{blue!23}{Fans Spin up When Computer Is Set to Sleep}& \colorbox{green!25}{Laptop Hardware},  \colorbox{green!25}{Hibernation}&4\\

\hline
\colorbox{blue!23}{Modern Standby}& \colorbox{blue!23}{Battery Draining While Asleep}& \colorbox{lime}{Computer Model}, \colorbox{green!25}{Power}&4\\
\hline
\colorbox{green!25}{Network Driver}\footnote{<vendor> PCIe FE / GBE / 2.5G / Gaming Ethernet network card drivers (v.NetAdapterCx)}& \colorbox{blue!23}{Problems with Modern Standby}& \colorbox{lime}{Laptop Brand and Model}, \colorbox{lime}{Windows Version}, \colorbox{green!25}{Automatic Driver Installation}&5\\
\hline
\colorbox{blue!23}{Modern Standby}&\colorbox{blue!23}{TextInputHost.exe Constantly Crashes}& \colorbox{lime}{Operating System},\colorbox{green!25}{RAM}, \colorbox{green!25}{Processor}&5\\

\hline

\end{tblr}
\caption{Causal variables extracted from "Modern Standby" feedback. The highlighted variables are \colorbox{blue!23}{pre-classified bugs/treatments}, variables that are \colorbox{lime}{mentioned in feedback/few-shot examples} and those that \colorbox{green!25}{are not explicitly mentioned in prompt.}}
\label{table:graph}
\begin{tblr}
{ 
colspec ={|p{16.1cm}|p{1.4cm}|},
row{1} = {lightgray},
}
\hline
\textbf{\small {Sequence of Events}}&\textbf{ \small{Heuristic 2}}\\
\hline
\colorbox{green!25}{Modern Standby -> sleep Mode -> TextInputHost.exe Crashes}
&3\\

\hline
\colorbox{green!25}{<vendor> PCIe FE / GBE / 2.5G / Gaming Ethernet Network Card Drivers (v.NetAdapterCx) -> Modern Standby -> System Reboots and Inability to Wake Up}\\
\colorbox{green!25}{Installing the Manufacturer's NDIS Version -> Proper Functioning of Modern Standby}
&3+2=5\\
\hline

 \colorbox{green!25}{Laptop Will Not Sleep Properly -> Fans Spin Up -> Forced to Hibernate}
&3\\

\hline
 \colorbox{blue!23}{Modern Standby -> Laptop Uses Power -> Fans Spin},
\colorbox{green!25}{Modern Standby -> changing closing Lid Behavior to Hibernate -> slower Wake-up Time}

&3+3=6\\
\hline

\colorbox{green!25}{Automatic sleep -> overheating -> Crash, keyboard Light ON -> Wake Up, Screen Not On -> Router Cannot Detect New Device Access, Fan Not Turning -> Body Very Hot}
&6\\
 \hline
\colorbox{green!25}{Waking Laptop From s0 Modern Standby Mode -> Task Scheduler Can Carry Out Backup Tasks}&2\\

\hline

 \colorbox{blue!23}{Modern Standby -> Laptop Overheats -> S3 Never Has That Problem}
 &3\\
\hline
\end{tblr}
\label{table:chain}
\caption{Chain of events extracted from "Modern Standby" feedback. \colorbox{blue!23}{Pre-classified} and \colorbox{green!25}{discovered} sequences are highlighted.}
\end{table*}

\section{Limitations}
In this work, we observed promising capabilities of LLMs in causal reasoning. However, there are still limitations that need to be addressed. These limitations include, but are not limited to:
\begin{itemize}
    \item In this project, we rely heavily on user feedback, while feedback received from users is highly prone to error and false assumptions due to lack of domain knowledge. Testing user's assumption  via causal pipeline is the most straightforward way to mitigate such errors.
    \item We proposed a greedy strategy to select confounders in the hope of increasing the likelihood of unconfoundedness. However, this assumption not only does not guaranty unconfoundedness, but may lead to incorrect selection of confounding factors as well. In order to avoid such situations, confounder selection strategies should be applied afterwards.
    \item Although LLMs uncover some important yet neglected factors, in many situations relevant telemetry data is not available. Nevertheless, identifying data gaps is as important as causal discovery.
\end{itemize}
\section{Conclusion}
In this work, we leverage LLMs and In-Context Learning to propose a self-consistent method for extracting causal insights from users' feedback. We propose a modified self-consistency technique which minimizes LLM hallucination and maximizes the pool of confounding variables. Our method extracts causal variables (i.e., treatment effect and confounders) as well as sequences of events that lead to the reported bugs.  We use these variables and sequences to define causal heuristics for scoring feedback based on their level of informativeness. For a given topic (e.g.,, "Modern Standby"), we observe that all extracted causal variables and sequences are in-domain insights which means our method is able to extract either previously known issues ($81\%$) or in-domain new bugs ($19\%$) while minimizes out-of-domain responses ($0\%$).
\newpage

\end{document}